\documentclass{article}
\usepackage{booktabs}

\usepackage{arxiv}

\usepackage[utf8]{inputenc} % allow utf-8 input
\usepackage[T1]{fontenc}    % use 8-bit T1 fonts
\usepackage{hyperref}       % hyperlinks
\usepackage{url}            % simple URL typesetting
\usepackage{booktabs}       % professional-quality tables
\usepackage{amsfonts}       % blackboard math symbols
\usepackage{nicefrac}       % compact symbols for 1/2, etc.
\usepackage{microtype}      % microtypography
\usepackage{lipsum}
\usepackage{graphicx}
\usepackage{comment}
\usepackage{amsmath}

\graphicspath{ {./images/} }

\title{Decoupling Representation and Learning in Genetic Programming: the LaSER Approach}

\author{
 Nam H. Le \\
  Vermont Complexity Center\\
  University of Vermont\\
  Burlington, VT 05405 \\
  \texttt{nam.le@uvm.edu} \\
   \And
 Josh Bongard \\
  Vermont Complexity Center\\
  University of Vermont\\
  Burlington, VT 05405 \\
  \texttt{jbongard@uvm.edu} \\
}

\begin{document}
\maketitle

\begin{abstract}

Genetic Programming (GP) has traditionally entangled the evolution of symbolic representations with their performance-based evaluation, often relying solely on raw fitness scores. This tight coupling makes GP solutions more fragile and prone to overfitting, reducing their ability to generalize. In this work, we propose \textbf{LaSER (Latent Semantic Representation Regression)} -- a general framework that decouples \emph{representation evolution} from \emph{lifetime learning}. At each generation, candidate programs produce features which are passed to an external learner to model the target task. This approach enables any function approximator, from linear models to neural networks, to serve as a lifetime learner, allowing expressive modeling beyond conventional symbolic forms.

Here we show for the first time that LaSER can outcompete standard GP and GP followed by linear regression when it employs non-linear methods to fit coefficients to GP-generated equations against complex data sets. Further, we explore how LaSER enables the emergence of innate representations, supporting long-standing hypotheses in evolutionary learning such as the Baldwin Effect. By separating the roles of representation and adaptation, LaSER offers a principled and extensible framework for symbolic regression and classification.

\end{abstract}

% keywords can be removed

\section{Introduction}

The remarkable success of modern machine learning, particularly deep learning, has been driven in part by the explicit separation of representation learning and task-specific optimization. 
In these systems, representation is encoded in the neural architecture, while learning is performed through gradient-based optimization of model parameters. 
This decoupling has proven effective in enhancing model flexibility, interpretability, and generalization across diverse tasks \cite{lecun2015deep, rumelhart1986learning, bengio2013representation}.

In contrast, GP traditionally evolves symbolic expressions whose fitness directly reflects predictive performance. 
This process entangles representation with learning, making it difficult to adapt or generalize expressions across different datasets or objectives \cite{koza1994genetic, Le2016}. 
As a result, GP solutions often become brittle, overfit to specific training instances, and lack the ability to fine-tune outputs to downstream tasks.

To address these limitations, we introduce \textbf{LaSER} (\textit{Latent Semantic Representation Regression}), a general-purpose framework that decouples representation learning from supervised task modeling in GP. 
Instead of evolving symbolic expressions that must directly fit target outputs, LaSER evolves programs that act as \emph{feature generators} -- producing intermediate representations that are passed to an external learner.
Formally, given a symbolic expression \( g(x) \) evolved by GP, LaSER composes this with a supervised model \( f \), such that the final prediction is given by \( f(g(x)) \). 
The learner \( f \) may be a linear model (e.g., ridge regression), a neural network, or any other regression or classification algorithm.
This decoupling enables more expressive and robust modeling, while preserving the interpretability and structure discovery strengths of GP.

Algorithmically, LaSER generalizes earlier ideas like Keijzer’s linear scaling~\cite{keijzer2003improving}, which applied a simple post-hoc transformation to evolved outputs. Unlike that work, LaSER decouples representation and learning entirely and supports arbitrary learners. Biologically, LaSER also draws inspiration from the \textit{Baldwin Effect} \cite{Baldwin:1896, Hinton:1987, Le2019}, which suggests that learned behaviors can guide evolutionary change, eventually becoming encoded in the genotype. 
In our case, the GP expression evolves to produce more useful representations, while the lifetime learner adapts its predictions based on those outputs. 
This interaction creates a pathway for learned adjustments to shape evolutionary trajectories.

\textbf{Our contributions are as follows:}
\begin{itemize}
    \item We present LaSER, a framework for explicitly separating symbolic representation and learning in GP.
    \item We show that LaSER generalizes Keijzer’s linear scaling by allowing any supervised learner to be used in the post-processing step.
    \item We conduct extensive experiments on symbolic regression benchmarks using a variety of learners (linear and nonlinear).
    \item We demonstrate that linear methods excel in polynomial domains, while nonlinear learners perform better in more complex, nonlinear tasks.
    \item We analyze how the Baldwin Effect emerges under LaSER, including evidence of reduced reliance on learning as evolution progresses.
\end{itemize}

The next section presents some related literature and how our LaSER pipeline can be drawn upon, before our experimental method and results.

\section{Related Work}
\label{sec:related-work}

Genetic Programming (GP) has long been applied to symbolic regression, where it evolves symbolic expressions that directly predict target outputs from raw input variables~\cite{koza1994genetic}. 
Standard GP evaluates candidate programs based on raw prediction error, tightly coupling the evolution of representation with its predictive fitness. 
While effective in simple scenarios, this direct coupling often results in fragile expressions that overfit training data and generalize poorly~\cite{Le2016, rosca1996generality}. 

In response, several enhancements have been proposed to improve GP’s robustness and expressiveness. 
These include modifying genetic operators to incorporate semantic awareness~\cite{moraglio2012geometric}, designing more selective pressure through methods such as lexicase selection~\cite{la2016epsilon}, or introducing new forms of expression construction, such as Geometric Semantic Programming (GSGP)~\cite{vanneschi2014geometric}. 
Although such methods have shown empirical benefits, they often lack modularity and remain tightly bound to GP’s core evolutionary loop. 
This integration makes them harder to adapt within broader machine learning workflows, and issues such as expression bloat remain persistent, particularly in semantic-based methods like GSGP \cite{vanneschi2014survey}.

In contrast, Keijzer's linear scaling~\cite{keijzer2003improving} departs from modifying the internal mechanics of GP. 
Rather than altering tree structures or genetic operators, it applies a post-hoc linear transformation -- typically via least-squares regression -- to the outputs of evolved programs. 
This introduces a simple, separate step to numerically rescale predictions, partially decoupling the evolved representation from the final prediction. Viewed differently, Keijzer’s approach distinguishes between the evolution of symbolic expressions and the minimization of error, hinting at a broader paradigm where loss minimization is applied independently from evolutionary search. 
The LaSER pipeline we shall be presenting in this paper embraces and extends this principle, treating loss minimization explicitly as a supervised learning process applied on top of evolved representations, and generalizing beyond linear scaling to accommodate arbitrary function approximators.

Modern machine learning systems, particularly deep learning architectures, have achieved remarkable success by learning intermediate representations that facilitate downstream tasks~\cite{bengio2013representation, lecun2015deep}. 
A central tenet in these systems is the decoupling of representation learning from task-specific modeling: lower layers of neural networks learn to extract semantically meaningful features, while upper layers perform prediction. 
This principle underpins major advances such as autoencoders for unsupervised representation learning~\cite{hinton2006reducing}, attention-based models like the Transformer~\cite{vaswani2017attention}, and large language models that pretrain general-purpose representations~\cite{radford2018improving}. 
It also enables transfer learning~\cite{weiss2016survey}, where learned representations can be reused across different tasks or domains. 

LaSER draws inspiration from this modularity: GP evolves symbolic representations, while an external learner maps these to predictions.  This separation facilitates flexible modeling, improves compatibility with modern pipelines, and enables analysis of representation evolution in symbolic domains.

Beyond engineering motivations, LaSER also connects to long-standing ideas in evolutionary biology. 
The \textit{Baldwin Effect}~\cite{Baldwin:1896, Hinton:1986, Le2019} suggests that learned behaviors can, over generations, become genetically assimilated — gradually encoded into innate traits through evolutionary pressure. 
This phenomenon has been studied in evolutionary computation like genetic algorithms ~\cite{whitley1994lamarckian, le2018adaptive}, and more in neuroevolution and neural architecture search where evolution can provide initial architectural representation before lifetime learning can fine tune the network \cite{floreano2008neuroevolution, NamAISB2019, Ackley+Littman:1992, fernando2018meta}

Recent work has explored incorporating components from modern machine learning into GP systems. For example, neural networks have been used to initialize or guide GP populations~\cite{Mundhenk2022GP}, hybrid approaches have embedded GP within deep learning frameworks~\cite{suganuma2017genetic}, and pretrained models have informed symbolic search~\cite{UnifiedPetersen2022}. These approaches aim to leverage the strengths of ML models -- such as differentiability, scalability, and data-driven representation learning -- and the interpretability of GP.

While promising in terms of empirical performance, these systems are largely engineering solutions. They typically mix heterogeneous components in an ad-hoc fashion, with little regard for conceptual coherence. As such, they deviate from the original vision of GP as a Darwinian process that evolves symbolic expressions through variation and selection ~\cite{koza1990genetic, koza1994genetic}. In particular, they fail to offer a principled framework in which learning occurs as a behavioral adjustment, without modifying the underlying program structure. This separation between learning and evolution is central to Baldwinian or Darwinian adaptation, yet remains largely absent from current symbolic regression frameworks.

LaSER provides a practical framework to examine this effect: by decoupling evolution and learning, we can analyze how evolved symbolic structures improve in raw (pre-learning) performance over generations, reflecting potential Baldwinian adaptation. While our focus is primarily algorithmic, this conceptual link highlights how LaSER bridges machine learning and biological learning perspectives in a unified framework.

In summary, while various prior efforts have introduced learning elements into GP, none provide a general framework that explicitly decouples the evolution of symbolic representations from downstream learning. 
Keijzer’s linear scaling offers an early instance of post-hoc adjustment, but it is limited to a fixed linear form and lacks conceptual modularity. 
Other hybrid approaches entangle representation and learning, restricting flexibility and making integration with modern machine learning techniques more difficult. 
\textbf{LaSER} fills this gap by providing a unified framework where GP evolves intermediate or latent symbolic features, and any supervised learner -- linear or nonlinear -- maps these features to predictions. 
This compositional abstraction not only improves performance but also enhances extensibility, positioning LaSER as a versatile tool for both symbolic regression and classification tasks.

\section{Experimental Method}

\subsection{Canonical GP Pipeline for Symbolic Regression}

\begin{figure}[htbp]
    \centering
    \includegraphics[width=0.95\linewidth]{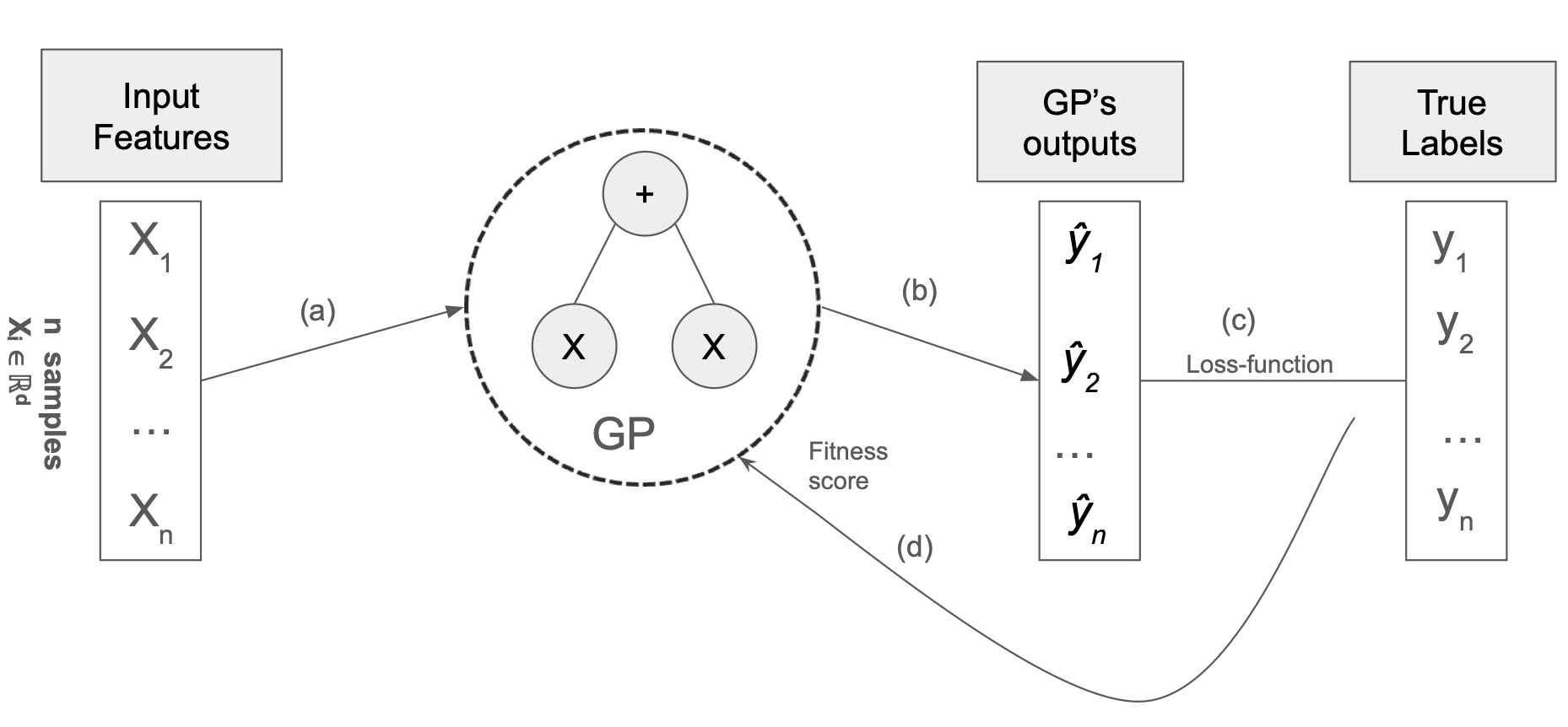}
    \caption{
        Canonical symbolic regression pipeline using GP.
        (a) Each individual is evaluated on the dataset to produce a semantic output vector. 
        (b) The output is compared directly to ground-truth labels using a loss function (e.g., MSE). 
        (c) The loss determines the individual's fitness used in selection.
    }
    \label{fig:gp_pipeline}
\end{figure}

Before introducing LaSER, we briefly review the standard symbolic regression pipeline using GP, shown in Figure~\ref{fig:gp_pipeline}. In canonical GP~\cite{koza1994genetic}, each individual in the population encodes a symbolic expression as a syntax tree. The goal is to evolve expressions that map input features to output targets.

Given a dataset of \( n \) samples, we represent the inputs as a matrix \( \mathbf{X} \in \mathbb{R}^{n \times d} \), where each row \( \mathbf{x}_i \in \mathbb{R}^d \) is a feature vector, and the targets as \( \mathbf{y} = [y_1, y_2, \dots, y_n] \in \mathbb{R}^n \).

Each GP individual defines a function \( f_{\text{GP}} \), which is evaluated on all input samples to produce a vector of predictions:
\[
\hat{\mathbf{y}} = f_{\text{GP}}(\mathbf{X}) = [f_{\text{GP}}(\mathbf{x}_1), \dots, f_{\text{GP}}(\mathbf{x}_n)]
\]
This prediction vector is also referred to as the \emph{semantic vector} of the individual.

A loss function \( \mathcal{L}(\hat{\mathbf{y}}, \mathbf{y}) \), typically the mean squared error (MSE), is used to compute the individual's fitness:
\[
\mathcal{L}(\hat{\mathbf{y}}, \mathbf{y}) = \frac{1}{n} \sum_{i=1}^{n} (\hat{y}_i - y_i)^2
\]

This fitness score guides the evolutionary process via selection and reproduction. However, since the GP tree must simultaneously discover both a useful internal representation and minimize the predictive error toward the target, this monolithic design often suffers from overfitting, bloated expressions, and reduced generalization.

To overcome these limitations, we propose a modular framework, LaSER, which explicitly separates the evolution of symbolic representations from the learning of predictive mappings, as in the following section.

\subsection{Latent Semantic Evolutionary Regression (LaSER) Pipeline}

LaSER extends the canonical GP pipeline by introducing an explicit separation between two roles: the evolution of symbolic \textit{representations}, and the learning of data-driven \textit{mappings} from those representations to target outputs. This decoupling enables greater flexibility in optimization and supports improved generalization through external learners.

Inspired by Baldwinian evolution, LaSER evaluates each individual not by its raw output alone, but by how well its induced \textit{semantic vector} supports downstream learning.

\begin{figure}[htbp]
    \centering
    \includegraphics[width=0.95\linewidth]{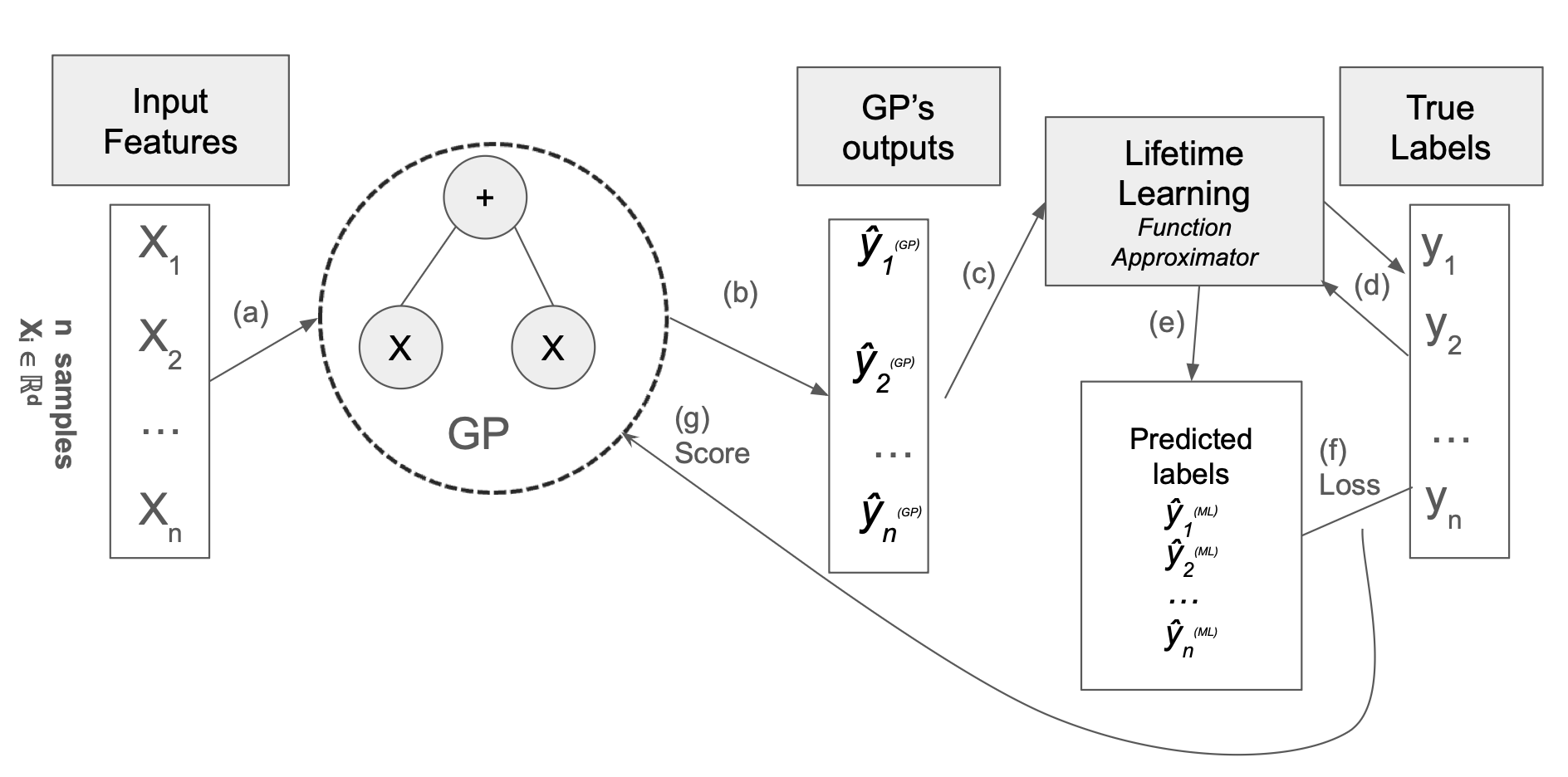}
    \caption{
    Overview of the LaSER pipeline. 
    (a) A GP individual encodes a symbolic function \( g(x) \) and is evaluated on the dataset to produce a semantic vector \( \hat{\mathbf{y}}^{(\text{GP})} \). 
    (b) This vector represents the symbolic behavior of the individual over the inputs. 
    (c) A supervised learner \( f \) is trained to predict the ground truth \( \mathbf{y} \) from the semantics. 
    (d--f) The learner's predictions \( \hat{\mathbf{y}}^{(\text{ML})} \) are compared to \( \mathbf{y} \) via a loss function. 
    (g) The resulting loss is used as the individual's fitness. 
    The symbolic tree remains unchanged, aligning with Baldwinian evolution.
    }
    \label{fig:laser_pipeline}
\end{figure}

As illustrated in Figure~\ref{fig:laser_pipeline}, each individual in the GP population encodes a symbolic function, denoted \( g(x) \). When evaluated on the dataset \( \mathbf{X} \in \mathbb{R}^{n \times d} \), it produces a semantic output vector:
\[
\hat{\mathbf{y}}^{(\text{GP})} = [g(\mathbf{x}_1), g(\mathbf{x}_2), \dots, g(\mathbf{x}_n)]^\top
\]

Rather than directly comparing \( \hat{\mathbf{y}}^{(\text{GP})} \) to the target vector \( \mathbf{y} \), LaSER introduces a \textbf{lifetime learning} stage. Here, a supervised model \( f \) is trained to map the GP semantics to the true outputs:
\[
\hat{\mathbf{y}}^{(\text{ML})} = f(\hat{\mathbf{y}}^{(\text{GP})})
\]

The learner \( f \) can be any parametric model -- linear regression, ridge, decision tree, neural network, gradient boosting, etc. -- allowing the system to handle both simple and highly nonlinear patterns. In classification tasks, \( f \) may be logistic regression or any probabilistic classifier.

The fitness assigned to the GP individual is then computed as a loss function comparing predictions to ground truth:
\[
\text{Fitness} = \mathcal{L}(\hat{\mathbf{y}}^{(\text{ML})}, \mathbf{y})
\]

Importantly, the GP expression \( g(x) \) remains fixed during learning; only the parameters of \( f \) are adjusted. This mirrors Baldwinian learning: an individual’s behavior improves through adaptation, but its genetic code (symbolic representation) is not directly modified by learning.

When \( f \) is a simple linear function (e.g., \( f(z) = \alpha z + \beta \)), the final model can be composed into a symbolic form. With more complex learners, LaSER produces a hybrid model: symbolic representations enhanced by data-driven adaptation.

Notably, when the lifetime learner \( f \) is restricted to a linear transformation (i.e., \( f(z) = \alpha z + \beta \)), LaSER reduces exactly to the linear scaling method proposed by Keijzer~\cite{keijzer2003improving}. Thus, Keijzer’s method can be interpreted as a special case within the broader LaSER framework.

\section{Experiments and Results}

\subsection{Evolutionary Setup}

All experiments were conducted using the \texttt{DEAP} evolutionary computation framework~\cite{fortin2012deap}. For both baseline GP and LaSER, the following hyperparameters were used unless otherwise specified:

\begin{itemize}
    \item \textbf{Population size}: 100
    \item \textbf{Generations}: 100
    \item \textbf{Crossover probability}: 0.8
    \item \textbf{Mutation probability}: 0.05
    \item \textbf{Selection method}: Tournament selection with size 3
    \item \textbf{Function set}: \{\texttt{+}, \texttt{-}, \texttt{*}, \texttt{/}, \texttt{sin}, \texttt{cos}, \texttt{exp}, \texttt{log}\}
    \item \textbf{Terminal set}: Input variables and ephemeral constants
    \item \textbf{Tree depth limits}: Max initial = 5; Max overall = 17
    \item \textbf{Elitism}: Top 1 individual preserved per generation
\end{itemize}

For each benchmark problem, we generate a dataset of 1000 input–output pairs, randomly split into 70\% training and 30\% test sets. All reported metrics are averaged over 30 independent runs with different random seeds to ensure statistical robustness.

\subsection{LaSER with Lifetime Learning Setup}

LaSER allows flexible integration of external learning models during fitness evaluation. To explore the impact of different lifetime learning strategies, we instantiate LaSER with a variety of supervised learners, ranging from linear models to more non-linear function approximators. This subsection details the configurations used for each learner.

\begin{itemize}
    \item \textbf{Linear Regression (LaSER-LR)}: A standard least-squares linear model with no regularization. This reproduces the behavior of Keijzer’s linear scaling~\cite{keijzer2003improving} in the LaSER framework.
    
    \item \textbf{Ridge Regression (LaSER-Ridge)}: A regularized linear model using $L_2$ penalty to prevent overfitting, controlled by default hyperparameters from \texttt{scikit-learn}.
    
    \item \textbf{Decision Tree (LaSER-Tree)}: A single tree regressor used to capture simple, axis-aligned non-linearities in the semantic space.
    
    \item \textbf{Random Forest (LaSER-RF)}: An ensemble of trees trained on bootstrapped semantic outputs, enabling better generalization across noisy representations.
    
    \item \textbf{Gradient Boosting (LaSER-GB)}: A stage-wise additive model combining weak learners, effective for modeling complex relationships within semantic outputs.
    
    \item \textbf{Multi-layer Perceptron (LaSER-MLP)}: A neural network with one hidden layer of 50 units, trained using early stopping and a maximum of 500 iterations. This model introduces highly flexible non-linearity while remaining compact enough to avoid overfitting.
\end{itemize}

After a GP individual produces its outputs over the training inputs, a lifetime learning step is applied to fit these outputs to the target values. Importantly, this learning is non-inheritable: the learned parameters are not passed to offspring or retained across generations. Consequently, modifying the choice of learner changes only the way fitness is computed -- without altering the evolutionary operators or the symbolic search process itself.

This modularity allows LaSER to be used as a plug-and-play framework to evaluate how different models benefit from symbolic feature evolution. In our experiments, we compare all learners on the same GP population to assess how learning capacity interacts with representation evolution.

\subsection{Nguyen Benchmarks: Revisiting Linear Scaling}

The Nguyen benchmark suite~\cite{mcdermott2012genetic} consists of 12 symbolic regression tasks with varying levels of algebraic complexity. These functions serve as a entry-level testbed for evaluating generalization in symbolic regression pipelines, as described in Table \ref{tab:nguyen_benchmarks} below.

\begin{table}[ht]
\centering
\begin{tabular}{lll}
\toprule
\textbf{Problem} & \textbf{Equation} & \textbf{Input Range} \\
\midrule
Nguyen-1  & $x^3 + x^2 + x$                             & $x \sim \mathcal{U}[-1, 1]$ \\
Nguyen-2  & $x^4 + x^3 + x^2 + x$                       & $x \sim \mathcal{U}[-1, 1]$ \\
Nguyen-3  & $x^5 + x^4 + x^3 + x^2 + x$                 & $x \sim \mathcal{U}[-1, 1]$ \\
Nguyen-4  & $x^6 + x^5 + x^4 + x^3 + x^2 + x$           & $x \sim \mathcal{U}[-1, 1]$ \\
Nguyen-5  & $\sin(x^2) \cdot \cos(x) - 1$               & $x \sim \mathcal{U}[-1, 1]$ \\
Nguyen-6  & $\sin(x) + \sin(x + x^2)$                   & $x \sim \mathcal{U}[-1, 1]$ \\
Nguyen-7  & $\log(x + 1) + \log(x^2 + 1)$               & $x \sim \mathcal{U}[-1, 1]$ \\
Nguyen-8  & $\sqrt{x}$                                  & $x \sim \mathcal{U}[0, 1]$  \\
Nguyen-9  & $\sin(x) + \sin(y^2)$                       & $x, y \sim \mathcal{U}[-1, 1]$ \\
Nguyen-10 & $2 \sin(x) \cos(y)$                         & $x, y \sim \mathcal{U}[-1, 1]$ \\
Nguyen-11 & $x \cdot y$                                 & $x, y \sim \mathcal{U}[-1, 1]$ \\
Nguyen-12 & $x^4 - x^3 + \frac{1}{2}y^2 - y$            & $x, y \sim \mathcal{U}[-1, 1]$ \\
\bottomrule
\end{tabular}
\caption{Nguyen benchmark functions used for symbolic regression evaluation.}
\label{tab:nguyen_benchmarks}
\end{table}

We begin by evaluating \textbf{LaSER-LR}, the variant of LaSER that applies linear regression as the lifetime learner. This configuration directly reproduces Keijzer’s linear scaling approach~\cite{keijzer2003improving}, where a linear model is fitted post hoc to the output of each GP individual. Within the LaSER framework, this becomes a modular baseline that isolates the impact of linear post-processing while preserving symbolic evolution dynamics.

To assess generalization performance, we compare test MSEs across 30 independent runs using the Wilcoxon signed-rank test. Table~\ref{tab:nguyen_mse_wilcoxon} summarizes the median test errors for each benchmark, including $p$-values and outcome direction. In nearly all cases, \textbf{LaSER-LR} outperforms standard GP, achieving significantly lower median errors. On Nguyen-8 and Nguyen-11, both methods achieve perfect or near-perfect fits, resulting in ties under the statistical test.

\begin{table}[ht]
\centering
\caption{Wilcoxon signed-rank test comparing test MSE medians of Standard GP and LaSER-LR on the Nguyen benchmarks. Results are based on 30 independent runs. “Tie” indicates no statistically significant difference or equal medians.}
\label{tab:nguyen_mse_wilcoxon}
\begin{tabular}{lrrrl}
\toprule
\textbf{Problem} & \textbf{p-value} & \textbf{GP Median} & \textbf{LaSER-LR Median} & \textbf{Direction} \\
\midrule
Nguyen-1 & 0.619467 & 0.000000 & 0.000237 & GP better \\
Nguyen-2 & 0.001383 & 0.005418 & 0.000426 & LaSER-LR better \\
Nguyen-3 & < 0.0001  & 0.023073 & 0.001077 & LaSER-LR better \\
Nguyen-4 & < 0.0001  & 0.038767 & 0.001478 & LaSER-LR better \\
Nguyen-5 & < 0.0001  & 0.002060 & 0.000000 & LaSER-LR better \\
Nguyen-6 & < 0.0001  & 0.006420 & 0.000217 & LaSER-LR better \\
Nguyen-7 & < 0.0001  & 0.033082 & 0.005012 & LaSER-LR better \\
Nguyen-8 & < 0.0001  & 0.000921 & 0.000023 & LaSER-LR better \\
Nguyen-9 & < 0.0001  & 0.029749 & 0.000132 & LaSER-LR better \\
Nguyen-10 & 0.000172 & 0.001439 & 0.000000 & LaSER-LR better \\
Nguyen-11 & 0.998375 & 0.000000 & 0.000000 & Tie \\
Nguyen-12 & < 0.0001  & 0.071468 & 0.012519 & LaSER-LR better \\
\bottomrule
\end{tabular}
\end{table}

We also evaluate predictive quality using the test $R^2$ metric, summarized in Table~\ref{tab:nguyen_r2_wilcoxon}. \textbf{LaSER-LR} consistently yields higher $R^2$ medians, indicating improved fit to test data. As with MSE, ties occur on Nguyen-11 where both methods perfectly model the target function. The results confirm that even simple post-processing via linear regression is sufficient to significantly boost performance in many symbolic regression tasks.

\begin{table}[ht]
\centering
\caption{Wilcoxon signed-rank test comparing test $R^2$ medians of Standard GP and LaSER-LR on the Nguyen benchmarks. Higher $R^2$ indicates better generalization.}
\label{tab:nguyen_r2_wilcoxon}
\begin{tabular}{lrrrl}
\toprule
\textbf{Problem} & \textbf{p-value} & \textbf{GP Median} & \textbf{LaSER-LR Median} & \textbf{Direction} \\
\midrule
Nguyen-1 & 0.545324 & 1.000000 & 0.999727 & GP better \\
Nguyen-2 & 0.001373 & 0.996397 & 0.999687 & LaSER-LR better \\
Nguyen-3 & < 0.0001  & 0.986872 & 0.999472 & LaSER-LR better \\
Nguyen-4 & < 0.0001  & 0.984529 & 0.999280 & LaSER-LR better \\
Nguyen-5 & < 0.0001  & 0.928719 & 0.999983 & LaSER-LR better \\
Nguyen-6 & < 0.0001  & 0.993798 & 0.999753 & LaSER-LR better \\
Nguyen-7 & < 0.0001  & 0.948501 & 0.993997 & LaSER-LR better \\
Nguyen-8 & < 0.0001  & 0.982427 & 0.999519 & LaSER-LR better \\
Nguyen-9 & < 0.0001  & 0.887671 & 0.999544 & LaSER-LR better \\
Nguyen-10 & 0.000296 & 0.998093 & 1.000000 & LaSER-LR better \\
Nguyen-11 & 0.089856 & 1.000000 & 1.000000 & Tie \\
Nguyen-12 & < 0.0001  & 0.844435 & 0.972480 & LaSER-LR better \\
\bottomrule
\end{tabular}
\end{table}

To further quantify robustness, we compute success rates based on the fraction of runs achieving $R^2 \geq 0.99$ on the test set (Table~\ref{tab:nguyen_success_rate}). \textbf{LaSER-LR} achieves perfect or near-perfect success on most benchmarks, including all of Nguyen-1 through Nguyen-6. On harder tasks like Nguyen-9 and Nguyen-12, standard GP shows a steep decline in reliability, while LaSER-LR maintains strong performance.

These results confirm that introducing a learning phas -- even with a simple linear model—substantially improves symbolic regression performance. \textbf{LaSER-LR} not only yields better generalization, but also significantly increases the stability and success rate of GP-based modeling.

\begin{table}[ht]
\centering
\caption{Success rates (\%) of LaSER-LR and Standard GP across the Nguyen benchmarks, based on 30 independent runs each. A run is considered successful if it achieves $R^2 \geq 0.99$ on the test set.}
\label{tab:nguyen_success_rate}
\begin{tabular}{lrr}
\toprule
\textbf{Benchmark} & \textbf{LaSER-LR} & \textbf{Standard GP} \\
\midrule
Nguyen-1  & 100.0 & 83.3 \\
Nguyen-2  & 100.0 & 66.7 \\
Nguyen-3  & 100.0 & 46.7 \\
Nguyen-4  & 96.7  & 23.3 \\
Nguyen-5  & 100.0 & 10.0 \\
Nguyen-6  & 100.0 & 56.7 \\
Nguyen-7  & 60.0  & 3.3  \\
Nguyen-8  & 96.7  & 33.3 \\
Nguyen-9  & 93.3  & 16.7 \\
Nguyen-10 & 93.3  & 70.0 \\
Nguyen-11 & 100.0 & 100.0 \\
Nguyen-12 & 26.7  & 0.0  \\
\bottomrule
\end{tabular}
\end{table}

\subsection{Comparing Lifetime Learners in LaSER}

While the previous section focused on LaSER using linear regression (LaSER-LR), the framework itself is agnostic to the choice of learning algorithm. To assess how different learning strategies affect symbolic regression performance, we evaluate LaSER using six lifetime learners: Linear Regression, Ridge Regression, Multi-Layer Perceptron (MLP), Decision Tree, Random Forest, and Gradient Boosting.

\begin{figure}[htbp]
    \centering
    \begin{tabular}{ccc}
        \includegraphics[width=0.3\textwidth]{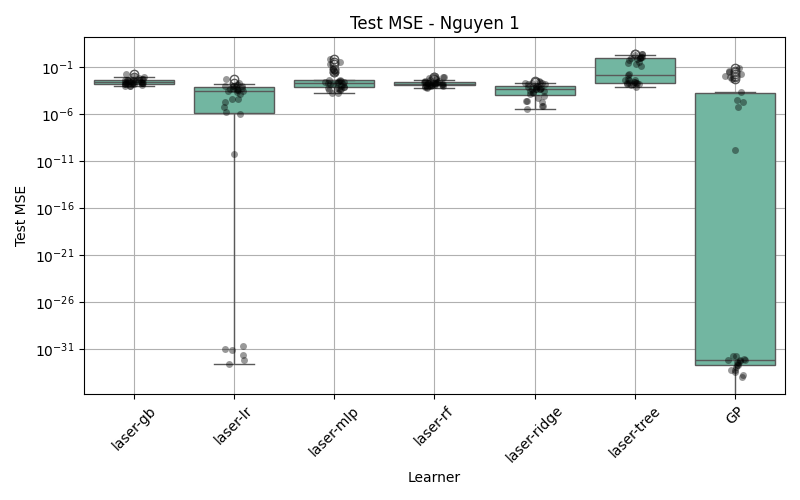} &
        \includegraphics[width=0.3\textwidth]{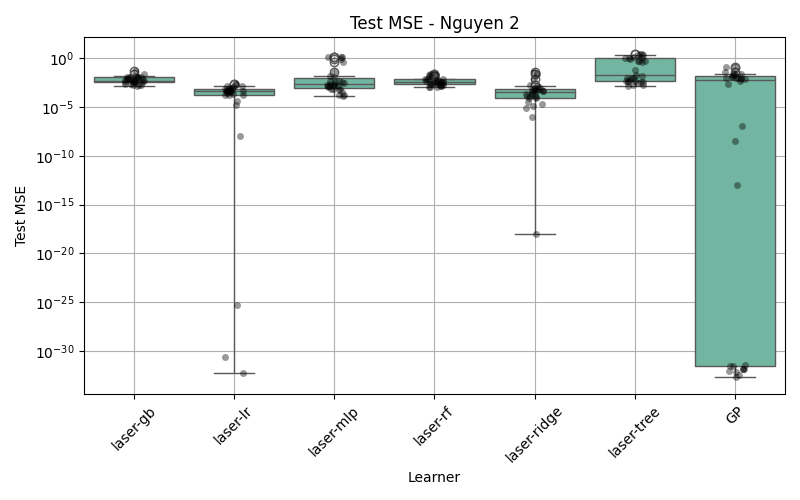} &
        \includegraphics[width=0.3\textwidth]{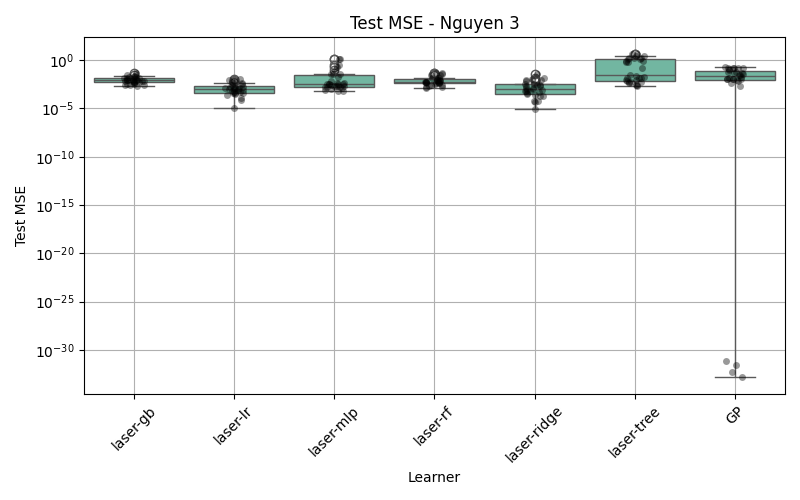} \\

        \includegraphics[width=0.3\textwidth]{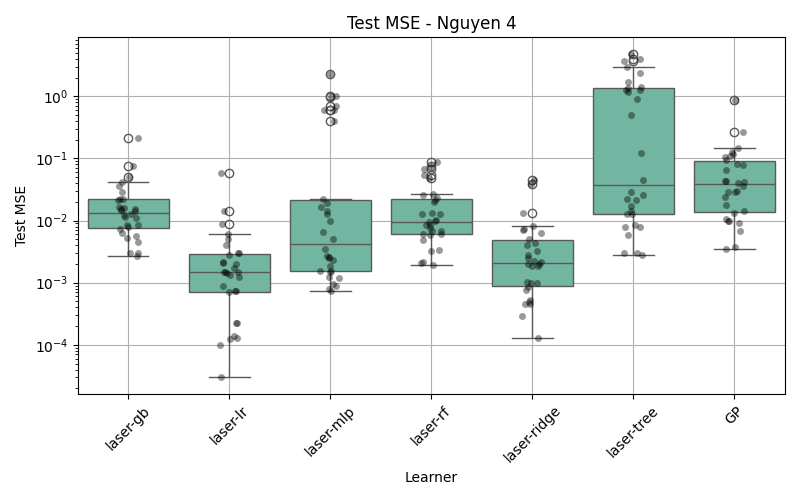} &
        \includegraphics[width=0.3\textwidth]{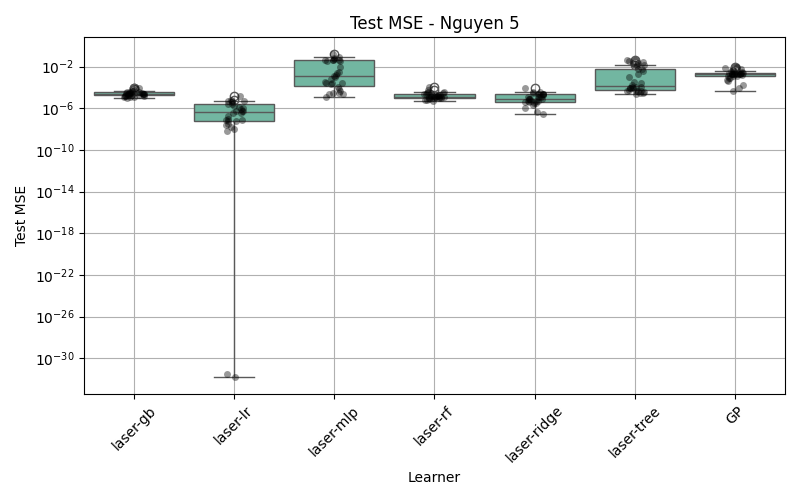} &
        \includegraphics[width=0.3\textwidth]{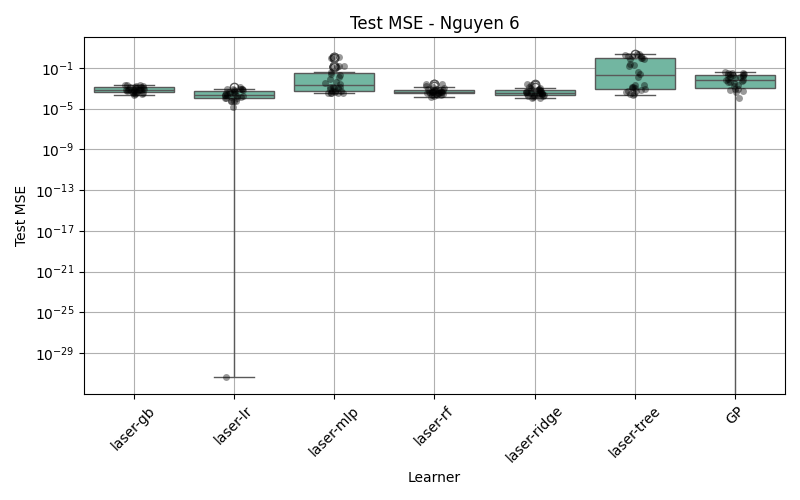} \\
       
        \includegraphics[width=0.3\textwidth]{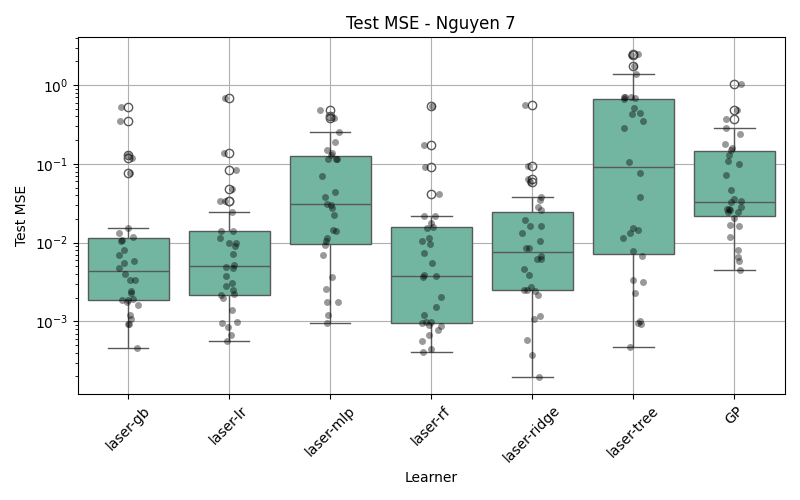} &
        \includegraphics[width=0.3\textwidth]{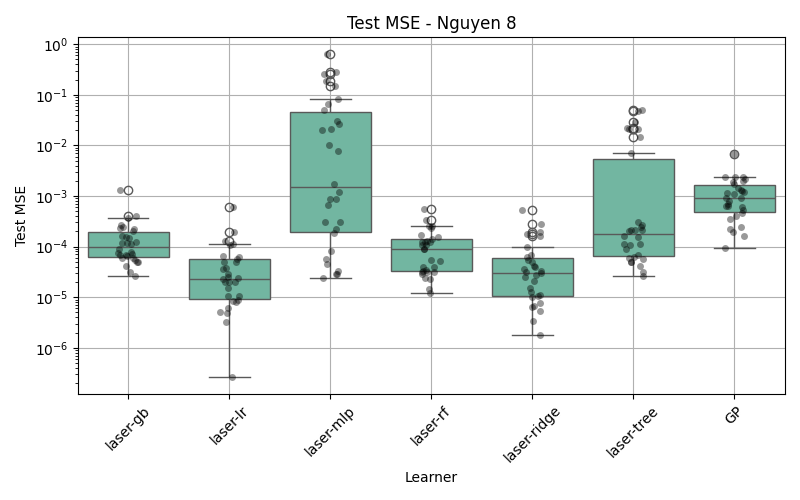} &
        \includegraphics[width=0.3\textwidth]{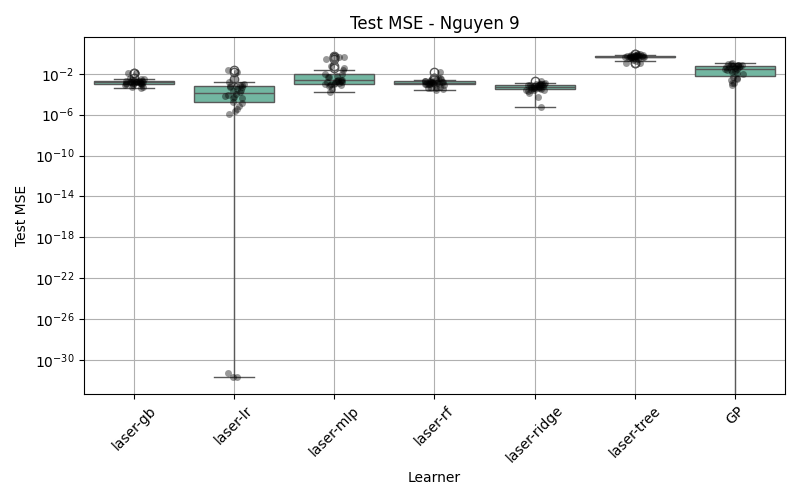} \\

        \includegraphics[width=0.3\textwidth]{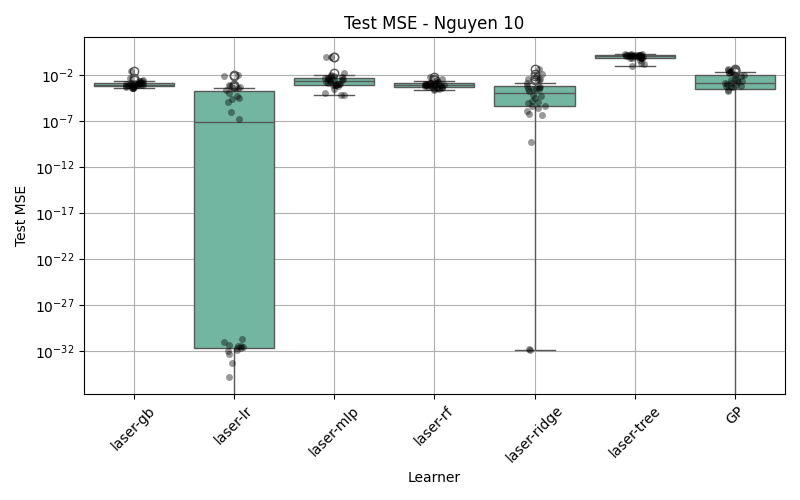} &
        \includegraphics[width=0.3\textwidth]{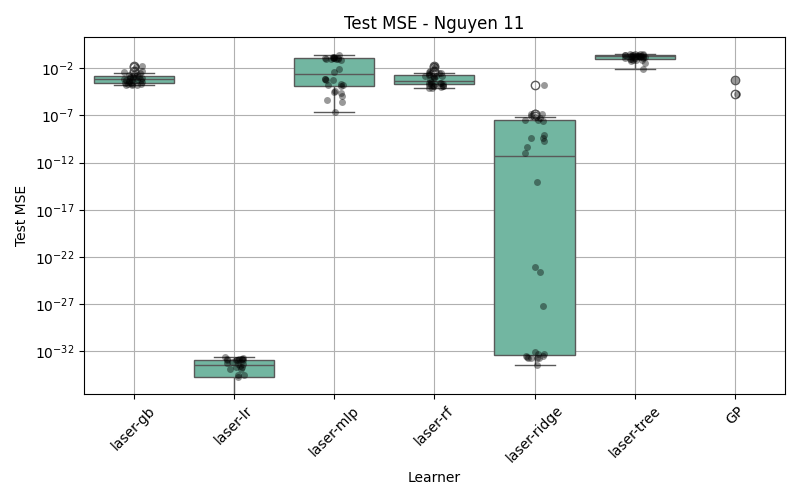} &
        \includegraphics[width=0.3\textwidth]{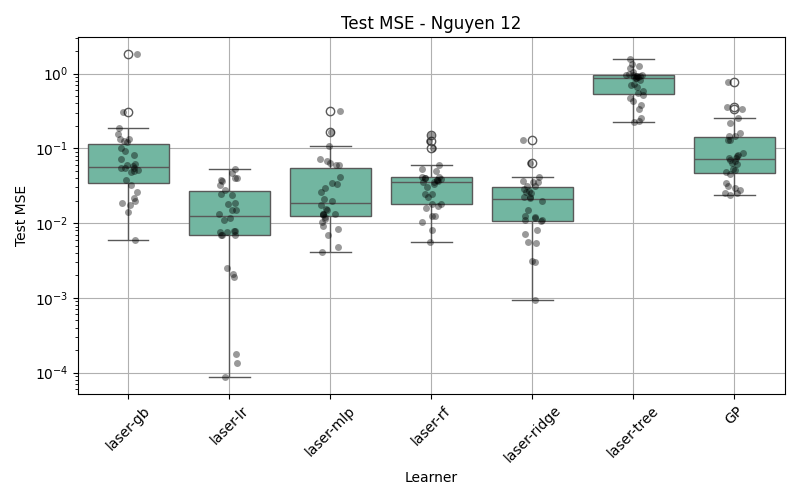} \\
    \end{tabular}
    \caption{Box plots of Test MSE (log scale) for LaSER and GP across the Nguyen benchmarks.}
    \label{fig:nguyen_mse_grid}
\end{figure}

\begin{table}[htbp]
\centering
\caption{Mean Test MSE for LaSER learners and GP across the Nguyen benchmark problems. Best (lowest) per row is highlighted in \textbf{bold}.}
\label{tab:nguyen_mse_comparison}
\begin{tabular}{lccccccc}
\toprule
Problem & GP & LaSER-GB & \textbf{LaSER-LR} & LaSER-MLP & LaSER-RF & LaSER-Ridge & LaSER-Tree \\
\midrule
nguyen1  & 0.0069 & 0.0038 & \textbf{0.0006} & 0.0496 & 0.0027 & 0.0008 & 0.4774 \\
nguyen2  & 0.0135 & 0.0077 & \textbf{0.0006} & 0.1658 & 0.0061 & 0.0025 & 0.5414 \\
nguyen3  & 0.0510 & 0.0110 & \textbf{0.0022} & 0.1124 & 0.0113 & 0.0037 & 0.8528 \\
nguyen4  & 0.0824 & 0.0238 & \textbf{0.0042} & 0.2226 & 0.0196 & 0.0056 & 0.9237 \\
nguyen5  & 0.0024 & \textbf{0.0000} & \textbf{0.0000} & 0.0222 & \textbf{0.0000} & \textbf{0.0000} & 0.0067 \\
nguyen6  & 0.0115 & 0.0009 & \textbf{0.0004} & 0.1322 & 0.0007 & 0.0006 & 0.5027 \\
nguyen7  & 0.1233 & 0.0439 & \textbf{0.0389} & 0.0940 & 0.0337 & 0.0345 & 0.4608 \\
nguyen8  & 0.0012 & 0.0002 & \textbf{0.0001} & 0.0616 & \textbf{0.0001} & \textbf{0.0001} & 0.0072 \\
nguyen9  & 0.0364 & 0.0023 & 0.0017 & 0.0646 & 0.0020 & \textbf{0.0006} & 0.4984 \\
nguyen10 & 0.0077 & 0.0022 & \textbf{0.0007} & 0.0645 & 0.0012 & 0.0030 & 1.2033 \\
nguyen11 & \textbf{0.0000} & 0.0021 & \textbf{0.0000} & 0.0554 & 0.0022 & \textbf{0.0000} & 0.1837 \\
nguyen12 & 0.1238 & 0.1349 & \textbf{0.0175} & 0.0429 & 0.0391 & 0.0238 & 0.7820 \\
\bottomrule
\end{tabular}
\end{table}

\begin{table}[htbp]
\centering
\caption{Mean Test $R^2$ scores for LaSER learners and GP across the Nguyen benchmark problems. Best (highest) per row is highlighted in \textbf{bold}.}
\label{tab:nguyen_r2_comparison}
\begin{tabular}{lccccccc}
\toprule
Problem & GP & LaSER-GB & \textbf{LaSER-LR} & LaSER-MLP & LaSER-RF & LaSER-Ridge & LaSER-Tree \\
\midrule
nguyen1  & 0.9930 & 0.9958 & \textbf{0.9994} & 0.9568 & 0.9972 & 0.9992 & 0.4495 \\
nguyen2  & 0.9862 & 0.9941 & \textbf{0.9995} & 0.8786 & 0.9951 & 0.9981 & 0.4722 \\
nguyen3  & 0.9720 & 0.9939 & \textbf{0.9986} & 0.9194 & 0.9940 & 0.9982 & 0.4368 \\
nguyen4  & 0.9653 & 0.9900 & \textbf{0.9980} & 0.8800 & 0.9909 & 0.9975 & 0.4584 \\
nguyen5  & 0.9031 & 0.9988 & \textbf{0.9999} & 0.1503 & 0.9992 & 0.9994 & 0.7476 \\
nguyen6  & 0.9863 & 0.9990 & \textbf{0.9996} & 0.8657 & 0.9992 & 0.9994 & 0.4335 \\
nguyen7  & 0.8815 & 0.9674 & 0.9730 & 0.8882 & \textbf{0.9791} & 0.9767 & 0.3876 \\
nguyen8  & 0.9769 & 0.9966 & \textbf{0.9990} & -0.0392 & 0.9980 & 0.9988 & 0.8707 \\
nguyen9  & 0.8851 & 0.9928 & 0.9951 & 0.8203 & 0.9944 & \textbf{0.9980} & -0.5554 \\
nguyen10 & 0.9891 & 0.9968 & \textbf{0.9989} & 0.9266 & 0.9984 & 0.9965 & -0.6117 \\
nguyen11 & 0.9998 & 0.9844 & \textbf{1.0000} & 0.5375 & 0.9854 & \textbf{1.0000} & -0.5728 \\
nguyen12 & 0.8040 & 0.8074 & \textbf{0.9694} & 0.9327 & 0.9314 & 0.9597 & -0.4093 \\
\bottomrule
\end{tabular}
\end{table}

Figure~\ref{fig:nguyen_mse_grid} and Tables~\ref{tab:nguyen_mse_comparison} and~\ref{tab:nguyen_r2_comparison} summarize the performance of LaSER across 12 classical symbolic regression benchmarks from the Nguyen suite. We report both box plots (MSE on log scale) and aggregate statistics (average Test MSE and $R^2$ over 30 runs per setting), comparing LaSER learners with vanilla Genetic Programming (GP). Best performing results are highlighted in bold.

Across the majority of benchmarks, LaSER-LR (linear scaling) stands out as the most effective learner. It consistently achieves the lowest error and highest $R^2$, particularly on smooth polynomial problems such as Nguyen-1 through Nguyen-6. This supports the long-standing observation that many symbolic regression targets in this suite are well approximated by linear combinations of evolved features -- making LaSER-LR especially well suited.

LaSER-Ridge (ridge regression) closely follows, offering similar robustness while mitigating overfitting in some settings. These findings reinforce the idea that linear models are often sufficient when GP already constructs expressive basis functions. The simplicity of these learners contributes to strong generalization, as seen in the tightly clustered box plots with minimal variance.

However, on structurally more complex benchmarks—such as Nguyen-7 (logarithmic) and Nguyen-8 (square root), we observe nonlinear learners like Random Forest (RF) and Gradient Boosting (GB) occasionally outperform linear models, albeit with greater variance. This suggests these learners can exploit residual patterns not captured by GP semantics alone.

In contrast, Decision Trees perform inconsistently and often degrade generalization, likely due to their high variance. MLPs also show unstable behavior, underperforming in several cases, potentially due to the small data regimes and sensitivity to initialization.

Overall, the results validate LaSER’s ensemble architecture and the strong inductive bias of LaSER-LR on analytic functions. As complexity grows, LaSER’s modular design allows seamless integration of more expressive learners, offering a compelling foundation for adaptive symbolic learning systems that scale with task difficulty.

\subsection{Performance on Complex Benchmarks: Vladislavleva Suite}

The Vladislavleva symbolic regression benchmarks~\cite{vladislavleva2008order} are designed to assess the ability of regression systems to handle increasingly complex functional forms, including multivariate, highly nonlinear, and non-polynomial structures. Each function introduces unique challenges in terms of variable interactions and landscape irregularity—offering a more demanding testbed than the Nguyen suite.

For reference, the suite includes the following eight target functions:

\begin{align}
f_1(x_1, x_2) &= \frac{e^{-(x_1 - 1)^2}}{1.2 + (x_2 - 2.5)^2} \\
f_2(x) &= e^{-x} x^3 \cos x \sin x \left( \cos x \sin^2 x - 1 \right) \\
f_3(x_1, x_2) &= f_2(x_1)(x_2 - 5) \\
f_4(x_1, x_2, x_3, x_4, x_5) &= \frac{10}{5 + \sum_{i=1}^5 (x_i - 3)^2} \\
f_5(x_1, x_2, x_3) &= 30 \cdot \frac{(x_1 - 1)(x_3 - 1)}{x_2^2 (x_1 - 10)} \\
f_6(x_1, x_2) &= 6 \sin x_1 \cos x_2 \\
f_7(x_1, x_2) &= (x_1 - 3)(x_2 - 3) + 2 \sin((x_1 - 4)(x_2 - 4)) \\
f_8(x_1, x_2) &= \frac{(x_1 - 3)^4 + (x_2 - 3)^3 - (x_2 - 3)}{(x_2 - 2)^4 + 10}
\end{align}

\begin{figure}[htbp]
    \centering
    \begin{tabular}{cc}
        \includegraphics[width=0.45\textwidth]{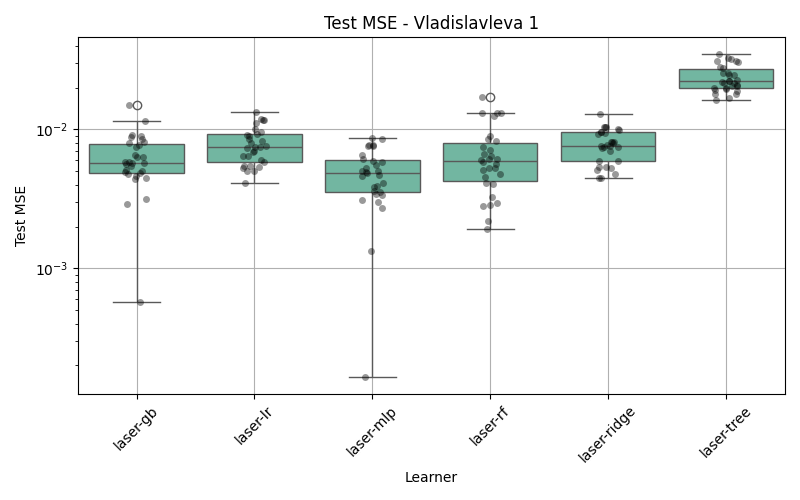} &
        \includegraphics[width=0.45\textwidth]{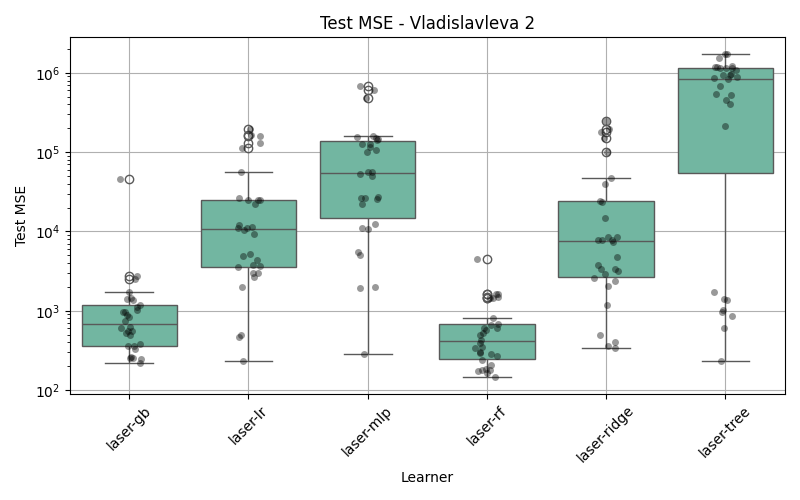} \\

        \includegraphics[width=0.45\textwidth]{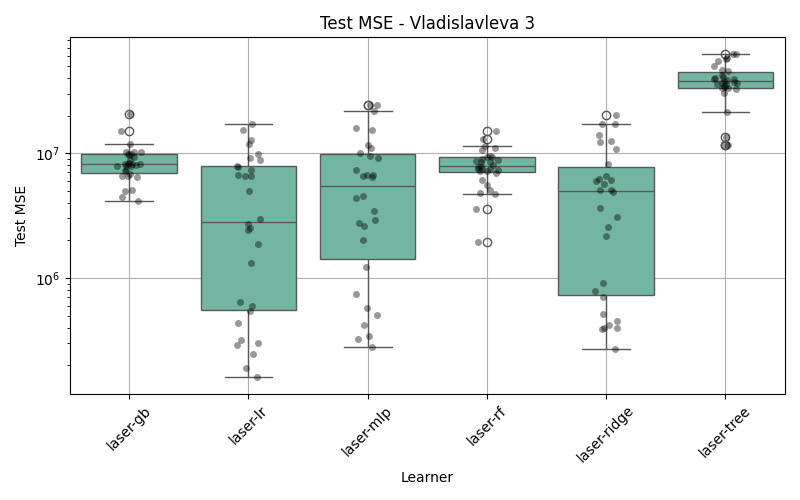} &
        \includegraphics[width=0.45\textwidth]{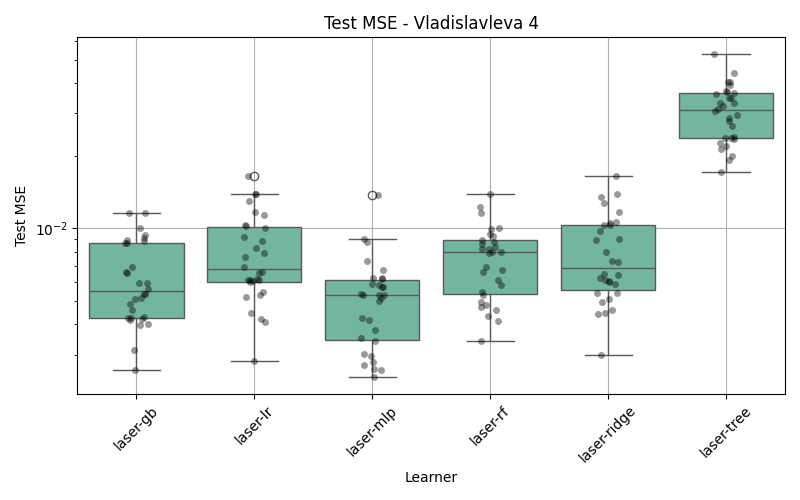} \\

        \includegraphics[width=0.45\textwidth]{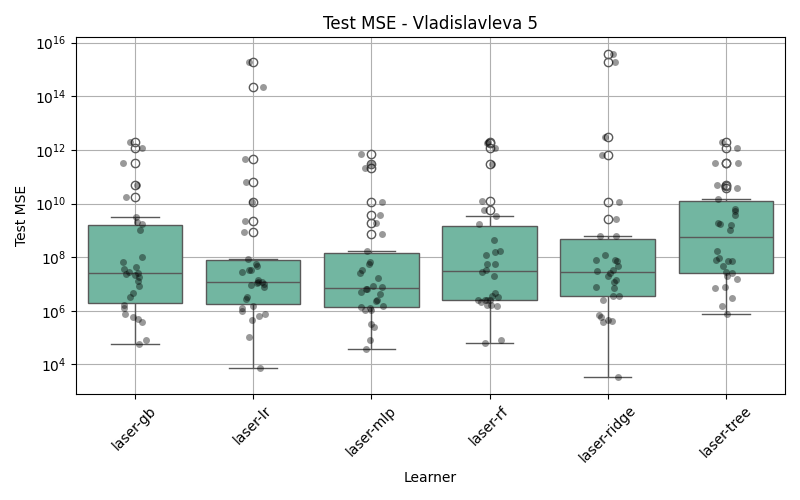} &
        \includegraphics[width=0.45\textwidth]{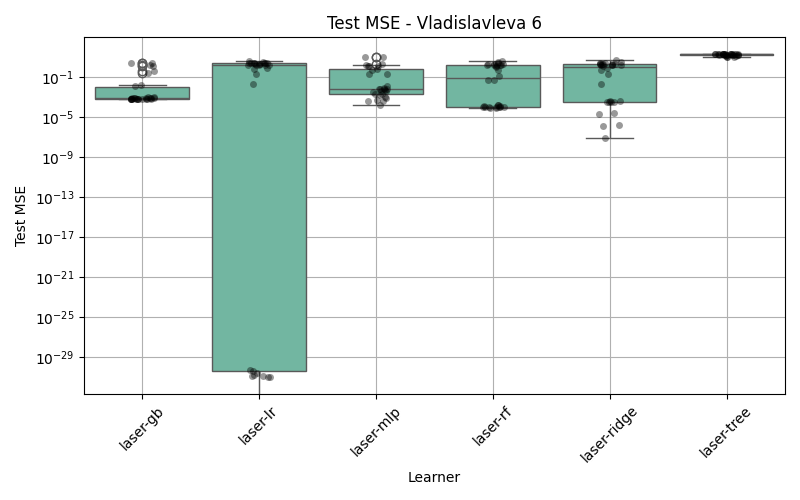} \\

        \includegraphics[width=0.45\textwidth]{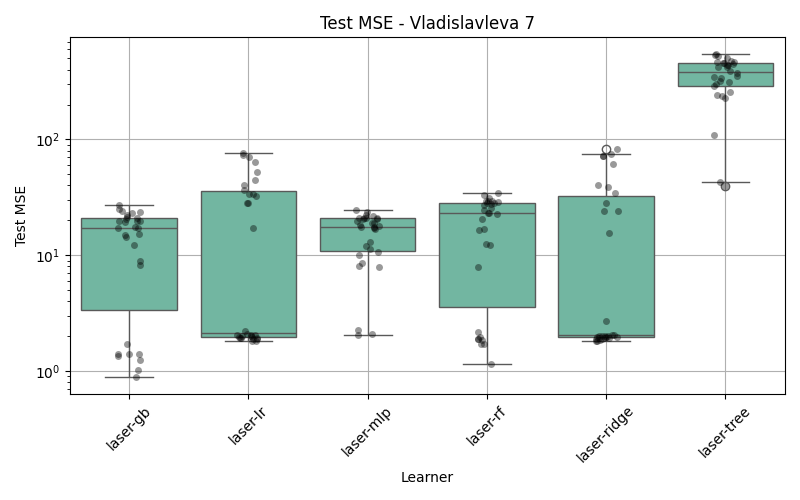} &
        \includegraphics[width=0.45\textwidth]{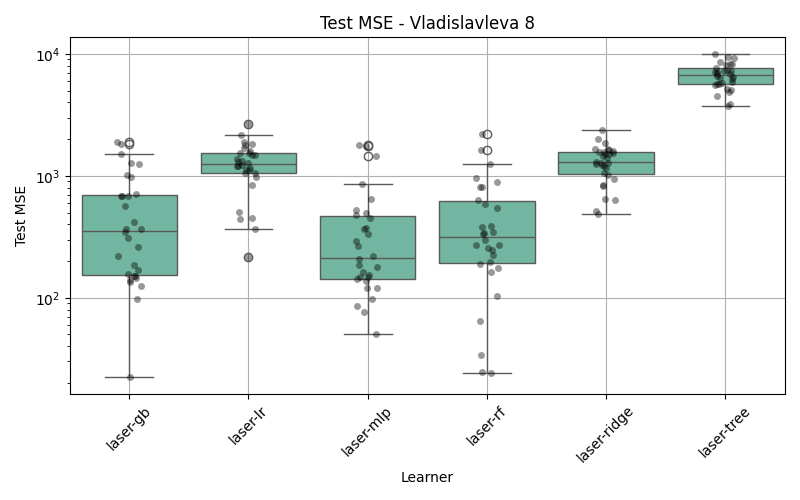} \\
        
    \end{tabular}
    \caption{Box plots (log scale) of Test MSE for LaSER variants across Vladislavleva-1 to Vladislavleva-8 benchmarks.}
    \label{fig:vla_mse_boxplots}
\end{figure}

While the Nguyen benchmarks showed that LaSER with linear lifetime learners (LaSER-LR and LaSER-Ridge) consistently delivered strong performance, those tasks predominantly involve smooth polynomial functions, which are easily captured by linear mappings. To explore whether this advantage persists under more difficult conditions, we turn to the Vladislavleva benchmarks -- known for their structural complexity and higher degrees of nonlinearity~\cite{vladislavleva2008order}.

Figure~\ref{fig:vla_mse_boxplots}, along with Table~\ref{tab:vla_mse_comparison} and Table~\ref{tab:vla_r2_comparison}, present a comparative analysis of LaSER learners on the Vladislavleva benchmarks. Each result reflects the distribution of test MSE (shown in log scale) and the average $R^2$ values across 30 independent runs.

Unlike the patterns observed in the Nguyen benchmarks, here we see a clear departure: \textbf{LaSER-LR} no longer consistently dominates. Instead, nonlinear learners such as \textbf{LaSER-MLP} and \textbf{LaSER-GB} emerge as top performers on several problems. For instance, LaSER-MLP achieves the lowest test error on Vla1 and Vla4, while LaSER-GB outperforms others on Vla6 and Vla7. This shift suggests that as the complexity of the target function increases—particularly with nonlinearity or high-dimensional interactions -- linear learners are insufficient to capture the necessary structure from evolved GP features.

Nevertheless, linear models remain competitive in certain problems (e.g., Vla3), and \textbf{LaSER-LR} continues to serve as a strong and efficient baseline. The broader takeaway lies in LaSER’s flexible architecture: its ability to seamlessly interchange the lifetime learning component without altering the GP evolutionary process. This modularity enables LaSER to tailor its learning strategy to the underlying problem structure.

As task complexity rises, the choice of learner becomes increasingly important. These results highlight LaSER’s adaptability -- not just in refining representations, but in aligning the inductive bias of the downstream model to the specific demands of symbolic regression tasks.

\begin{table}[ht]
\centering
\caption{Test MSE (lower is better) for LaSER using different lifetime learners on the Vladislavleva benchmarks. Best values are highlighted in \textbf{red}.}
\label{tab:vla_mse_comparison}
\begin{tabular}{lrrrrrr}
\toprule
\textbf{Problem} & \textbf{LaSER-GB} & \textbf{LaSER-LR} & \textbf{LaSER-MLP} & \textbf{LaSER-RF} & \textbf{LaSER-Ridge} & \textbf{LaSER-Tree} \\
\midrule
vla1  & 0.0064 & 0.0079 & \textbf{0.0049} & 0.0067 & 0.0078 & 0.0237 \\
vla2  & 2364.9350 & 34815.3404 & 116462.5437 & \textbf{708.9041} & 36779.3022 & 714494.7647 \\
vla3  & 8589712.0259 & \textbf{5003149.8821} & 7120068.0702 & 7973793.0464 & 5824657.3038 & 38399686.2531 \\
vla4  & 0.0063 & 0.0080 & \textbf{0.0052} & 0.0075 & 0.0080 & 0.0308 \\
vla5  & 1.2055e+11 & \textbf{7.1799e+13} & 4.1022e+10 & 1.7843e+12 & 1.9445e+10 & 1.3332e+13 \\
vla6  & \textbf{0.2727} & 1.2615 & 0.9312 & 0.8512 & 1.0591 & 16.2742 \\
vla7  & \textbf{14.1113} & 22.1229 & 15.5895 & 18.2749 & 20.1347 & 360.2344 \\
vla8  & 562.7317 & 1263.0675 & \textbf{409.9133} & 488.0160 & 1299.7373 & 6686.2380 \\
\bottomrule
\end{tabular}
\end{table}

\begin{table}[ht]
\centering
\caption{Test $R^2$ (higher is better) for LaSER using different lifetime learners on the Vladislavleva benchmarks. Best values are highlighted in \textbf{red}.}
\label{tab:vla_r2_comparison}
\begin{tabular}{lrrrrrr}
\toprule
\textbf{Problem} & \textbf{LaSER-GB} & \textbf{LaSER-LR} & \textbf{LaSER-MLP} & \textbf{LaSER-RF} & \textbf{LaSER-Ridge} & \textbf{LaSER-Tree} \\
\midrule
vla1  & 0.4997 & 0.3900 & \textbf{0.6103} & 0.4670 & 0.3889 & -0.8796 \\
vla2  & 0.9966 & 0.9478 & 0.8171 & \textbf{0.9990} & 0.9478 & -0.0810 \\
vla3  & 0.5963 & \textbf{0.7688} & 0.6780 & 0.6288 & 0.7382 & -0.7823 \\
vla4  & 0.5870 & 0.5081 & \textbf{0.6722} & 0.5099 & 0.5083 & -0.9906 \\
vla5  & -26.8876 & -4.3384e+06 & \textbf{-6.6991} & -5.4148e+03 & -4.3391e+06 & -1673.4895 \\
vla6  & \textbf{0.9686} & 0.8562 & 0.8932 & 0.9027 & 0.8808 & -0.8435 \\
vla7  & \textbf{0.9361} & 0.8983 & 0.9286 & 0.9182 & 0.9068 & -0.6273 \\
vla8  & 0.8506 & 0.6566 & \textbf{0.8846} & 0.8682 & 0.6484 & -0.8059 \\
\bottomrule
\end{tabular}
\end{table}

\subsection{Is There Evidence of the Baldwin Effect?}

The Baldwin Effect describes how learned behaviors can, over evolutionary time, become encoded directly into the genotype. In the context of LaSER, this would mean that as evolution proceeds, individuals improve their raw (pre-learning) predictions, reducing dependence on the learning mechanism. In the most extreme case, learning becomes redundant -- yielding what we call \emph{perfect innateness}.

We define a solution as \textbf{perfectly innate} if:
\begin{itemize}
    \item The raw output of the GP individual achieves \textbf{zero error} (MSE = 0) on both training and test sets.
    \item The learned linear model becomes an identity map: \( \alpha = 1 \), \( \beta = 0 \).
\end{itemize}

In such cases, the final output simplifies to:
\[
\hat{y}(x) = \alpha \cdot f_{\text{GP}}(x) + \beta = f_{\text{GP}}(x),
\]
indicating that the evolved symbolic representation itself captures the target function exactly.

To assess this phenomenon, we perform a retrospective analysis using LaSER with linear regression (LaSER-LR) across the 12 Nguyen benchmarks. These tasks are known to be relatively easy for symbolic regression methods, and frequently admit exact solutions with zero generalization error -- making them ideal for detecting potential Baldwinian assimilation.

Table~\ref{tab:baldwin_summary} shows, for each benchmark, the number of runs (out of 30) in which perfect innateness was reached at any generation.

\begin{table}[ht]
\centering
\caption{Number of runs (out of 30) in which LaSER-LR achieved \textbf{perfect innateness} (zero MSE and \( \alpha = 1, \beta = 0 \)).}
\label{tab:baldwin_summary}
\begin{tabular}{lr}
\toprule
\textbf{Benchmark} & \textbf{\# Runs with Perfect Innateness} \\
\midrule
Nguyen-1  & 21/30 \\
Nguyen-2  & 22/30 \\
Nguyen-3  & 10/30 \\
Nguyen-4  & 10/30 \\
Nguyen-5  & 1/30  \\
Nguyen-6  & 6/30  \\
Nguyen-7  & 5/30  \\
Nguyen-8  & 30/30 \\
Nguyen-9  & 13/30 \\
Nguyen-10 & 0/30  \\
Nguyen-11 & 20/30 \\
Nguyen-12 & 0/30  \\
\bottomrule
\end{tabular}
\end{table}

These results show that in many cases, the correct behavior does become fully encoded in the evolved symbolic expression, making the learning step superfluous. However, this effect is benchmark-dependent: it is most prevalent on simpler, univariate problems (e.g., Nguyen-1, -2, -8), and largely absent from more complex or multivariate ones (e.g., Nguyen-10, -12).

While a full investigation of the Baldwin Effect would require tracking fitness dynamics of raw versus learned models across generations, this early analysis suggests that LaSER provides a viable platform to study such phenomena. Future work will explore how different learning mechanisms, task types, and semantic architectures impact the emergence of innate solutions.

\section{Discussion and Future Work}

This work introduced LaSER, a modular framework that decouples the evolution of representations from the downstream learning task. By doing so, LaSER enables greater flexibility in selecting learning algorithms without altering the symbolic search process. On the Nguyen benchmark suite, we showed that LaSER with a simple linear learner (LaSER-LR) performs remarkably well -- often outperforming standard GP and other LaSER variants. This replicates and extends the classical linear scaling method proposed by Keijzer, now interpreted as a special case within a broader compositional framework. However, when applied to the more complex Vladislavleva benchmarks, LaSER-LR no longer dominates. Instead, nonlinear learners such as MLP and Gradient Boosting achieve better generalization on several tasks, demonstrating the importance of selecting an appropriate inductive bias for the learning component.
We also presented preliminary evidence for a form of the Baldwin Effect within LaSER: in many runs, evolution discovers representations that achieve perfect accuracy without requiring any additional learning, a phenomenon we refer to as ``perfect innateness.'' Taken together, these findings highlight LaSER’s strength as a flexible and principled framework for integrating symbolic evolution with modern machine learning, while also opening new avenues for analyzing evolutionary learning dynamics.

Despite these promising results, our current evaluation has limitations. First, the majority of our analysis focused on relatively low-dimensional regression benchmarks, particularly the Nguyen suite. While these tasks are widely used and well -- understood, they may not fully capture the challenges of real-world symbolic modeling, such as noise, redundancy, or large-scale feature interactions. Second, although LaSER supports arbitrary learning algorithms, we primarily explored standard regressors with minimal hyperparameter tuning. More sophisticated learners (e.g., neural architectures or kernel-based models) could potentially yield even stronger performance, but would require careful integration to preserve interpretability and computational feasibility.

Future work can extend LaSER along several promising axes. One direction is to explore richer model classes as learners, such as symbolic neural networks or sparse kernel machines, to better capture highly nonlinear or discontinuous mappings. Another avenue involves dynamically adapting the learning strategy during evolution -- for instance, by evolving or selecting the appropriate learner per individual or generation. Beyond predictive performance, LaSER also opens up new possibilities for studying evolutionary learning dynamics, such as characterizing the conditions under which learned behaviors become internalized over generations. ALso, applying LaSER to domains beyond symbolic regression -- such as classification task ((e.g., with logistic regression as the learner), program synthesis, control, or scientific discovery \cite{bongard2007automated} -- may reveal further benefits of its compositional design.

Beyond the single-vector setting explored in this paper, LaSER naturally extends to richer, population-level semantics. Instead of treating each GP individual as a standalone feature generator, we can view the entire population as producing a matrix of latent representations—one semantic vector per individual. This ensemble-style perspective aligns with the Pittsburgh-style paradigm in evolutionary computation~\cite{smith1980learning}. A separate supervised learner can then be trained to map this matrix of features to the target outputs, enabling more expressive modeling. Depending on the task, this learner can be linear or nonlinear, ranging from kernel methods to deep networks. This variant reframes symbolic regression as a standard supervised learning problem, where the evolved population defines a learned feature space. Such extensions are especially promising for problems involving high-dimensional inputs or complex mappings, where single-vector semantics may be insufficient. We leave the exploration of these directions to future work.

On the theoretical side, LaSER invites deeper analysis of its fitness dynamics and generalization behavior. For example, ablation studies could assess the individual contribution of the learning step, while bias–variance decomposition or PAC-style analysis \cite{valiant1984theory} could help characterize its learning capacity under different conditions.

Together, these directions position LaSER not just as a practical method for symbolic regression, but as a general framework for uniting evolutionary computation with representation learning in a principled and extensible manner.

\bibliographystyle{unsrt}  
\bibliography{references} 

\end{document}